%% file: root.tex
\documentclass[letterpaper, 10 pt, journal, twoside]{ieeeconf}  % ICRA format

\IEEEoverridecommandlockouts                              % This command is only
\let\labelindent\relax

\input{preamble.tex}

\input{glossary.tex}

\input{notation.tex}

\let\oldtwocolumn\twocolumn \renewcommand\twocolumn[1][]{%
    \oldtwocolumn[{#1}{
    \begin{center}
           \includegraphics[width=\textwidth]{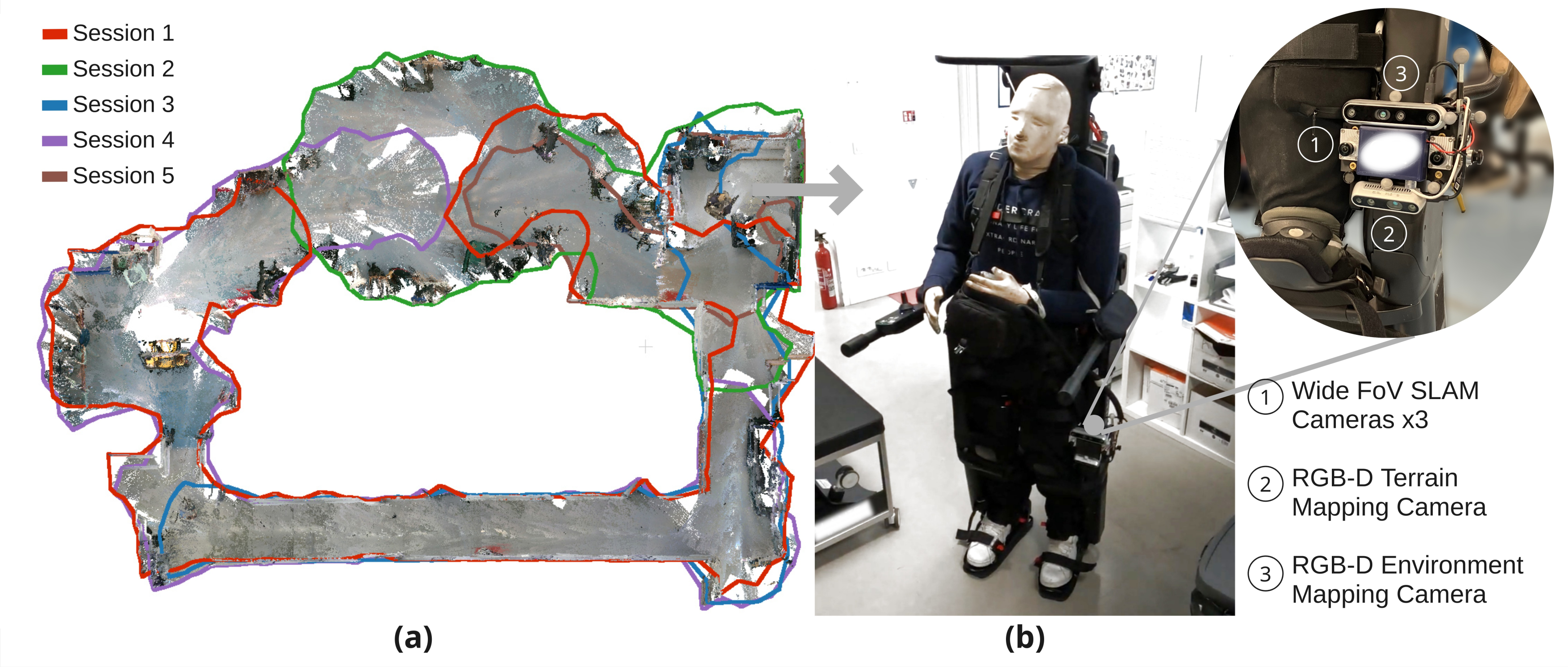}
           \captionof{figure}{LT-Exosense is capable of merging multiple sessions generated by a previous work, Exosense, a vision-centric scene understanding system with its sensing unit (\textbf{Top-Right}) integrated into a self-balancing exoskeleton (\textbf{b}). The merged map (\textbf{a}) contains five sessions with colored contours indicating the coverage area of each session. Such a merged map can be further converted into a navigation map, enabling obstacle-free planning spanning multiple sessions.}
           \label{fig:cover}
        \end{center}
    }] }

\begin{document}
\title{ \LARGE \bf LT-Exosense: A Vision-centric Multi-session Mapping System for Lifelong Safe Navigation of Exoskeletons}

\author{Jianeng Wang, Matias Mattamala, Christina Kassab, Nived Chebrolu, Guillaume Burger, Fabio Elnecave, Marine Petriaux, Maurice Fallon}

\maketitle

\begin{abstract}
Self-balancing exoskeletons offer a promising mobility solution for individuals with lower-limb disabilities. For reliable long-term operation, these exoskeletons require a perception system that is effective in changing environments. In this work, we introduce \textbf{LT-Exosense}, a vision-centric, multi-session mapping system designed to support long-term (semi)-autonomous navigation for exoskeleton users.
LT-Exosense extends single-session mapping capabilities by incrementally fusing spatial knowledge across multiple sessions, detecting environmental changes, and updating a persistent global map. This representation enables intelligent path planning, which can adapt to newly observed obstacles and can recover previous routes when obstructions are removed. 
We validate LT-Exosense through several real-world experiments, demonstrating a scalable multi-session map that achieves an average point-to-point error below \SI{5}{\centi\meter} when compared to ground-truth laser scans. We also illustrate the potential application of adaptive path planning in dynamically changing indoor environments.
\end{abstract}

\begin{keywords}
Multi-session Mapping, Wearable Robotics, Prosthetics and Exoskeletons, RGB-D Perception, Mapping
\end{keywords}

\section{Introduction}
Self-balancing exoskeletons provide a transformative solution enabling mobility-impaired individuals to walk independently---offering an alternative to wheelchairs and crutches. The main efforts have focused on human-compliant hardware design and control strategies~\cite{Tricomi2025Leveraging}. However, the deployment of self-balancing exoskeletons still remains confined to structured clinical and therapeutic contexts~\cite{tian2024exo}. This limits the benefit that these systems can bring to patients' daily lives and their long-term rehabilitation.

Beyond the clinical setting, the daily real-world usage of exoskeletons offers the immense potential of increased independence for their users while also reducing the occurrence of secondary health conditions, which improves the quality of life for patients with lower-limb disability \cite{miller2016clinical, van2024improvement}. However, achieving this goal requires not only advances in hardware and control algorithms but also effective perception to enable safe operation in dynamic environments. A persistent, long-term global map is essential for this. The map should enable the exoskeleton to plan long-distance paths between different rooms or floors, while also using local path-planning to follow trajectories and to avoid obstacles. This assistive semi-autonomy is helpful for stroke patients with both lower- and upper-limb impairments who may be unable to use a joystick precisely and would benefit from an exoskeleton safety layer ~\cite{lora2023robotic}. The ability to detect and respond to environment changes is therefore critical for safe operation.

A recently published system called Exosense~\cite{wang2025exosense} introduced a vision-based scene understanding method for exoskeletons that can generate detailed home-scale scene representations. The sensing unit was designed to be rigidly attached to the upper leg of an exoskeleton, so as to avoid sensor occlusion by the user’s body. This setup however introduces a jerky walking motion pattern, making accurate motion estimation difficult. While Exosense presents a system for exoskeleton localization and navigation, it could only operate for a single session and had no capacity to accumulate environmental knowledge over extended periods of time or to respond to spatial changes across successive mapping sessions.

To improve upon these limitations, we present \emph{LT-Exosense}, a change-aware, multi-session mapping system tailored for the long-term deployment of self-balancing exoskeletons in real-world environments. LT-Exosense can integrate spatial data from multiple exploration sessions to incrementally build a persistent map. It can detect and track environment changes, and update the global map to reflect the latest state of the world. This capability facilitates lifelong, intelligent navigation by allowing the exoskeleton to reuse maps of previously explored areas. It can also adapt to new conditions, and plan safe paths to familiar destinations. LT-Exosense has the potential to improve the experience of an exoskeleton user by providing intuitive mobility assistance.

The main contributions of our work are:
\begin{itemize}
\item LT-Exosense, a multi-session mapping system for self-balancing exoskeletons that captures terrain traversability information as well as identifies changes and obstacles in realistic dynamic environments. 
\item We validate LT-Exosense's ability to identify object-level change and to carry out multi-session reconstruction in a busy office environment.
\item We demonstrate a real-world adaptive path planning pipeline that can re-route around detected obstacles using the updated multi-session map.
\end{itemize}

\section{Related Work}\label{sec:related_work}
\subsection{Multi-session Visual SLAM}\label{sec:multi_session_mapping_review}
Traditional SLAM algorithms estimate a robot’s trajectory while simultaneously constructing a map of its environment, which makes them a key building block for autonomous systems. However, most conventional SLAM systems support only a single, continuous exploration session. In contrast, \emph{multi-session SLAM}~\cite{Schmid2025} goes further and supports long-term and large-scale operations by incrementally fusing the outputs of multiple SLAM sessions---whether performed by a single robot across different time intervals or by a team of robots collaboratively. This capability enables persistent mapping, robust localization when revisiting a place, and resilience to environmental changes over time.

In multi-session visual SLAM, the system must recognize previously visited places across different sessions using visual inputs. This is inherently challenging due to changes in lighting conditions, viewpoint, and scene appearance. Labbe and Michaud \cite{labbe2022multi} present a multi-session visual SLAM framework centered around re-localization, with each individual session built using RTAB-Map \cite{Labbe2019RTAB}. Their work evaluates various visual descriptors for illumination-invariant place recognition and loop closure. Experimental results indicate that learning-based feature detectors and matchers (e.g., SuperPoint \cite{DeTone2018superpoint} and SuperGlue \cite{sarlin20superglue}) offer improved robustness to appearance changes, albeit at the cost of increased computation and memory. To mitigate this, the framework incorporates a graph reduction strategy \cite{labbe2018long} to conserve resources while maintaining localization accuracy.

Dedicated multi-session mapping systems like \emph{maplab} \cite{schneider2018maplab} provide a tightly integrated pipeline for vision-based SLAM. It uses ROVIO \cite{Bloesch2015ROVIO} to construct individual sessions, saving pose graphs, keyframes, image features, and associated resources for inter-session place recognition, merging, and reconstruction. The updated system, \emph{maplab 2.0} \cite{cramariuc2022maplab}, expands support to heterogeneous sensor modalities and robot platforms, becoming agnostic to odometry sources. It also supports storing non-visual data (e.g., LiDAR scans, GPS), enabling more versatile graph optimization constraints for tasks like multi-agent mapping and semantic mapping. This system was successfully deployed in  DARPA Subterranean Challenge \cite{tranzatto2022cerberus} to support collaborative mapping and navigation of aerial and legged robots. While maplab itself does not target exoskeletons, which can be viewed as a class of legged robotic systems, its design motivates the type of adaptability we seek in the LT-Exosense system for multi-session exoskeleton mapping.

LT-Exosense adapts multi-session SLAM techniques for assistive mobility, focusing on long-term usability for self-balancing exoskeletons rather than general-purpose mapping. Our system achieves reliable map fusion across multiple sessions. It also incorporates change detection in order to support navigation in evolving environments, thereby bridging the gap between SLAM research and real-world deployment on an exoskeleton system.

\subsection{Change Detection}\label{sec:change_detection_review}
As mapping research matures, there is an increasing demand for long-term autonomy in dynamic environments for which the ability to detect changes in the scene over time is essential. Change detection enables robots to adapt their behavior in response to environmental variations and it is widely used in applications such as environment monitoring \cite{Pretto2021Building}, infrastructure inspection \cite{staniaszek2024autoinspect}, and disaster response \cite{Ohno2010Trials}. These changes may range from highly dynamic (e.g., pedestrians and vehicles) to semi-static alterations that evolve over longer periods.

Changes in a scene can be determined via geometric analysis of map representations. Grid-based structures such as elevation maps \cite{Fankhauser2018Elevation} and OctoMap \cite{Hornung2013Octomap} support ray-tracing techniques that update occupancy based on sensor ray traversal. By continuously integrating sensor measurements, the representation can adapt to changes, but without explicitly modeling the change. Although accurate, these methods are computationally intensive due to the need to process every cell along each ray. Real-time deployment often requires hardware acceleration, such as a GPU \cite{Miki2022Elevation}.

Visibility reasoning simplifies the change detection problem by checking whether a newly visible point is occluded by a previously observed point. If true, the previous point is classified as a change \cite{Pomerleau2014Long-term}. While this simplifies the problem, such methods are sensitive to incidence angle ambiguity---especially on ground surfaces---leading to misclassifications \cite{Lim2021ERASOR}. To mitigate this, visibility is often encoded as an auxiliary feature in downstream classifiers \cite{Kim2020Removert}.

Volumetric maps such as Signed Distance Fields (SDFs) and occupancy grids allow online change detection by modeling free space. Systems like Dynablox \cite{schmid2023dynablox} and DUFOMap \cite{daniel2024dufomap} detect changes when new sensor measurements violate prior free-space assumptions. For inter-session analysis, approaches like LiSTA \cite{rowell2024lista} and BeautyMap \cite{Jia2024BeautyMap} align volumetric maps and perform voxel-level differencing to detect environmental changes. These approaches typically assume that the compared maps are spatially complete and densely observed, and primarily target LiDAR sensors, which have a wide field-of-view coverage.

LT-Exosense adapts the volumetric change detection approach of LiSTA \cite{rowell2024lista} to support incremental mapping with RGB-D sensing focusing on the exoskeleton use case. Furthermore, it handles the issue of non-overlapping areas between the different sessions and maintains a single lifelong map to reflect the latest environment state to support practical downstream tasks such as adaptive path planning. This unlocks the potential for future exoskeletons to navigate dynamic environments safely and efficiently over time.

\section{System}\label{sec:system}
\begin{figure*}
  \centering
  \includegraphics[width=\linewidth]{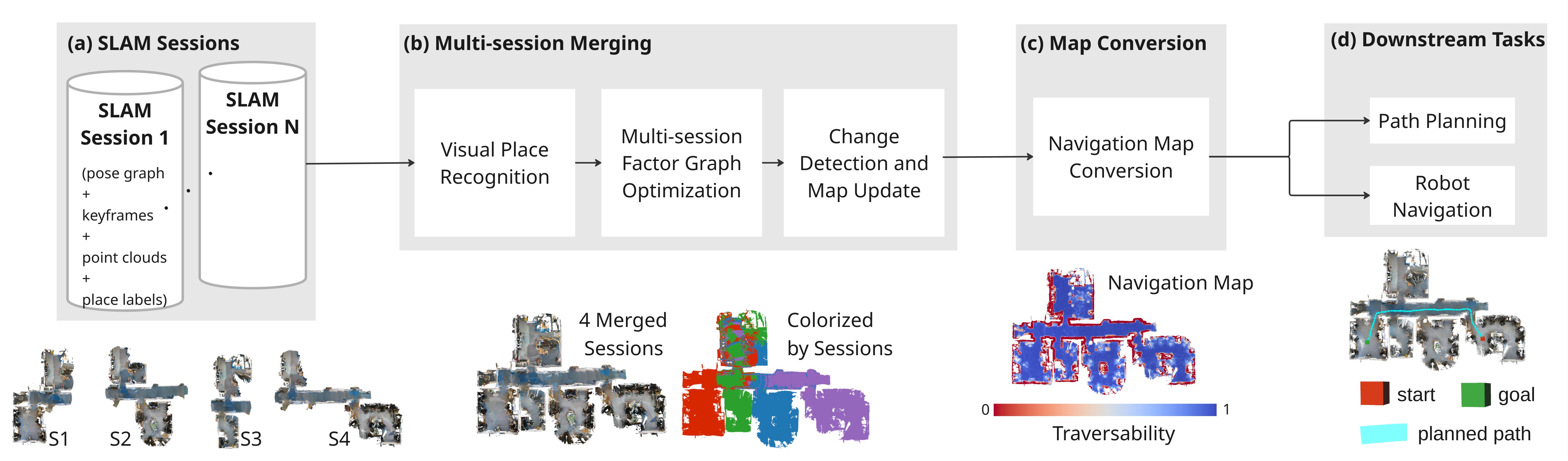}
  \caption{Overview of the LT-Exosense system. Multiple SLAM sessions with keyframe images and point cloud submaps (\textbf{a}) are registered into one common reference frame, where change detection is performed to reflect the latest environment changes for safe navigation of exoskeletons (\textbf{b}). The merged map can be further converted into elevation maps with traversability estimates (\textbf{c}). Obstacle-free walkable paths can be planned on the elevation maps (\textbf{d}).}
  \label{fig:system_overview}
\end{figure*}

The overall architecture of the LT-Exosense system is presented in \figref{fig:system_overview}. Our system receives multiple SLAM maps captured over time as input, each one produced from tracking cameras and RGB-D cameras data (\secref{sec:single_session_creation}). The SLAM maps from multiple sessions are aligned and merged to form a unified map, where environment changes are detected and updated to reflect the latest state of the environment (\secref{sec:multi_session_merging}). The unified map is then converted into an elevation map-based representation encoding features such as terrain geometry, semantics, and traversability (\secref{sec:navigation_map_conversion}). These features can in turn support downstream navigation tasks of the exoskeleton (\secref{sec:downstream_task}).

\subsection{Single-session Map Creation}\label{sec:single_session_creation}
To create a single SLAM map, we adapt the mapping pipeline from the \emph{Exosense} scene understanding system \cite{wang2025exosense}. This approach represents the environment as a collection of interconnected \textbf{submaps} associated to a pose graph. This structure allows the entire map to be globally optimized by adjusting the poses of these submaps whenever a loop closure is detected.

More specifically, each SLAM map consists of a pose graph of vertices and edges, along with associated resources linked to each vertex. A vertex represents the $SE(3)$ pose represented in the session's map frame, $\mathbf{T}_{\text{m}, \text{b}}$, where $\text{m}$ is the fixed map frame and $\text{b}$ is the robot base---represented in that session map frame. The edges of the graph are derived from either relative odometry (consecutive vertices), or loop closures (non-consecutive, when the robot revisits the same place). To each vertex we associate different data or representations---namely the stereo image pair, a point cloud submap and a semantic place label describing the vertex, such as the type of the room. These resources are later used to merge multiple sessions and to perform change detection to create a lifelong map.

\subsection{Multi-session Merging}\label{sec:multi_session_merging}
\subsubsection{Visual Place Recognition}
For every new SLAM session map, we perform visual place recognition using the images associated with a pose graph vertex by computing a global descriptor for each query image (from base pose $b_q$) and matching it against existing session maps in the database to propose potential loop closures (at base pose $b_p$).
The global descriptors consist of visual bag-of-words descriptors from ORB \cite{Rublee2011ORB} features and are  matched using DBoW \cite{dbow}. We use these descriptors for both intra- and inter-session loop closure proposals.

\subsubsection{Multi-session Factor Graph Optimization}
The visual place recognition module returns multiple potential loop closures for each query image. For each potential match between a query image (at base frame $\text{b}_q$) and a candidate image from a prior session (at base frame $\text{b}_p$), we compute the relative transformation $\hat{\mathbf{T}}_{\text{b}_p, \text{b}_q}$. This is achieved by finding SIFT \cite{lowe2004distinctive} feature matches and then using the Perspective-n-Point (PnP) method within a RANSAC scheme \cite{Fischler1981}.

Among those candidates, the match with the highest number of inliers is selected. The corresponding relative pose is then added as an edge constraint to the factor graph. If multiple prior sessions exist, a query image from the new session may match images in multiple sessions. This introduces inter-session constraints that link the new session to multiple existing sessions which promotes better global alignment and consistency in the merged map. The cost function considering the full inter-session matched set $\mathcal{M}$ is written as:
\begin{equation}
\mathcal{J}_{\text{inter}} = \sum_{(b_p, b_q) \in \mathcal{M}} 
\left\| 
\ln \left( 
\hat{\mathbf{T}}_{\text{b}_p, \text{b}_q}^{-1} \cdot 
\left( \mathbf{T}_{\text{m}, \text{b}_p}^{-1} \cdot \mathbf{T}_{\text{m}, \text{b}_q} \right) 
\right)^{\vee}\right\|_\Sigma^2,
\end{equation}
where $b_p$ and $b_q$ is the inter-session matched pose pair and $\Sigma$ is the covariance matrix associated with this relative pose estimate (from PnP registration).

Once all vertex-associated images are processed, we select the map frame of the first session to anchor all the remaining sessions, and then build a full merged graph by connecting pose graphs from individual sessions using the inter-session edges. The merged graph is then optimized using the Levenberg-Marquardt algorithm \cite{nocedal1999numerical} with a Cauchy loss function \cite{Lee2013Cauchy} to produce a globally consistent map for all the sessions.

To manage memory efficiency, each session (point clouds, images, descriptors) is stored locally in the file system. During multi-session merging, only the pose graphs are initially loaded into memory. The data-intensive resources of the graphs (e.g., images, point clouds) are only accessed on demand during relevant operations. This effectively minimizes peak memory usage.

\subsubsection{Change Detection for Latest Map Update} % to reflect latest merged map
As the robot incrementally explores its environment, the environment may experience change (typically furniture moving around). To maintain an up-to-date representation while preserving unchanged regions, we adopt the volumetric differencing approach presented in LiSTA \cite{rowell2024lista}, which incrementally updates the merged map as each new session is integrated.

Each merged session, comprising optimized poses and point cloud submaps, is converted into an octree representation by OctoMap~\cite {Hornung2013Octomap} that partitions the mapped space into occupied and free voxels. 
We then define a prior map and its corresponding octree $\mathcal{O}_{\text{p}}$, representing the current global map state prior to merging a new session. For the octree of a new session, $\mathcal{O}_{\text{c}}$, we perform a differencing operation between $\mathcal{O}_{\text{p}}$ and $\mathcal{O}_{\text{c}}$ to identify spatial change. This results in a removed octree $\mathcal{O}_{\text{r}}$, containing the occupied voxels present in $\mathcal{O}_{\text{p}}$ but absent in $\mathcal{O}_{\text{c}}$, and an added octree $\mathcal{O}_{\text{a}}$, with newly occupied voxels in $\mathcal{O}_{\text{c}}$ that were previously free in $\mathcal{O}_{\text{p}}$. Additionally, we compute the change-free prior octree, $\tilde{\mathcal{O}}_{\text{p}}$, by subtracting the removed nodes from the prior octree to isolate the unchanged structure:
\begin{equation}
\mathcal{\tilde{O}}_{\text{p}} = \mathcal{O}_{\text{p}} - \mathcal{O}_{\text{r}} = \mathcal{O}_{\text{p}} - ( \mathcal{O}_{\text{p}} \ominus \mathcal{O}_{\text{c}}),
\end{equation}
where $\mathcal{A} - \mathcal{B}$ denotes node deletion, and $\mathcal{A} \ominus \mathcal{B}$ denotes octree differencing. These operations return a set of voxels in the overlapping region with these different occupancy states.

The updated octree is produced by combining it with the current session’s octree:
\begin{equation}
\mathcal{O}_{\text{l}} = \mathcal{\tilde{O}}_{\text{p}}  + \mathcal{O}_{\text{c}},
\end{equation}
where $\mathcal{A} + \mathcal{B}$ merges two octrees in the same frame, with occupied nodes thereby overwriting free nodes.

This volumetric differencing strategy not only captures meaningful object-level changes but also removes misaligned or inconsistent point cloud data from prior sessions. As a result, the final merged map remains geometrically coherent and suitable for downstream navigation.

\subsection{Navigation Map Conversion}\label{sec:navigation_map_conversion}
\subsubsection{Point cloud map to Elevation map}
With the latest global map updated through multi-session merging and change detection, we prepare it for robot navigation by converting the point cloud submaps into a set of consistent local elevation maps. The resulting collection of elevation maps encodes the terrain geometry and traversability information required for downstream planning.

We first cluster spatially adjacent vertices in the merged pose graph that share the same place label. For each cluster, we compute the 3D bounding volume of all associated point cloud submaps. This bounding volume is then used to crop the corresponding region from the global map. 

Before converting the cropped point cloud to its final map representation, we need to first remove non-terrain points.
Overhanging structures, such as ceilings, can corrupt the resulting elevation map by introducing spurious height values that do not correspond to walkable terrain. To mitigate these dangling points, we introduce a coarse-to-fine filter to remove them.
For each point cloud submap, we partition the space into a 2D grid aligned with the x-y plane. Within each grid cell, we cluster nearby points based on their heights and keep only the lowest cluster. This process is repeated over several iterations using progressively finer grid resolutions.

After removing dangling points, the point clouds are converted into elevation maps using the method by Jelavic et al.~\cite{jelavic2021towards}. Each point cloud is projected onto a 2D grid at a predefined resolution, and the height of each cell is computed as the mean $z$-value of all points falling within it.

\subsubsection{Traversability Analysis}
To demonstrate the utility of LT-Exosense for exoskeleton navigation,  we present a prototype implementation for traversability analysis as one potential application of the generated maps, designed to show their suitability for downstream tasks like path planning. It is important to note that this analysis pipeline has not been integrated into a closed-loop control system with the exoskeleton.

For each cell $i$ in the elevation map, we define a traversability score, $t_i \in [0,1]$, representing how difficult it would be for the exoskeleton to step onto it, where $t_i=1$ corresponds to fully traversable and $t_i=0$ to an untraversable cell.
To compute the terrain traversability, we use the same approach in \cite{wang2025exosense}, by first selecting the neighborhood $\mathcal{C}_i$ of cell $i$ as the set of all cells within a nominal maximum stride length $s^{\ast}$ of the robot. We then compute the maximum elevation difference within this neighborhood as
\begin{equation}
h_i^{\text{max}} = \max{(|h_j - h_i|)}, \hspace{5mm} j\in \mathcal{C}_i.
\end{equation}
Given the maximum height height $h^{\ast}$ the exoskeleton can step on, the traversability score of a cell is
\begin{equation}
t_i = 1 - \min{(\frac{h_i^{\text{max}}}{h^{\ast}},1)}.
\end{equation}
This score serves as a conservative estimate of how safely the exoskeleton can navigate from the current cell to its neighbors, given its locomotion capabilities.

\subsection{Path Planning}\label{sec:downstream_task}
The resultant elevation maps are merged to form a unified representation of the environment. A global probabilistic roadmap (PRM) \cite{Kavraki1996PRM}  is then computed on top of this merged map for geometric motion planning. The PRM is built by randomly sampling a set of nodes representing valid robot configurations across the traversable regions of the merged elevation map. Nodes are connected if a path between them is determined to be collision-free. This process results in a graph that approximates the connectivity of the free space for the robot's safe navigation.

By setting the start and goal poses on the map, the PRM can be queried to connect them to nearby nodes in the existing graph. The resulting sequence of nodes constitutes a feasible geometric path from the start to the goal.

\def\thesubsectiondis{Exp \Alph{subsection}.} 
\renewcommand{\thesubsection}{Exp \Alph{subsection}}

\section{Experiments}
\label{sec:experiment}

We conducted a series of experiments to evaluate the performance of LT-Exosense in four areas: SLAM trajectory alignment accuracy (\ref{sec:exp_trajectory}), change detection performance (\ref{sec:exp_change}), multi-session mapping quality (\ref{sec:exp_mapping}), and its applicability to exoskeleton navigation tasks (\ref{sec:exp_planning}).

We first use the EuRoC dataset~\cite{Burri2016EuRoC} to assess our multi-session SLAM trajectory alignment.
We also used a custom-collected dataset that includes multiple sessions with varying environmental conditions, recorded with a multi-camera device from Exosense \cite{wang2025exosense} mounted on a person's or exoskeleton's leg. 

We denote each dataset by $D^{d}_{a}$, where $d$ indicates the day the dataset was collected and $a$ indicates the area. Sequences from the same day contain no environment change, while those recorded on different days include object-level changes.

 \vspace{0.25em}
 \noindent\textbf{(H) Human.} This dataset includes four sequences recorded with the Exosense sensing unit mounted on a person's thigh. Two same-day sequences, $H^{d_1}_{a_1}$ and $H^{d_1}_{a_2}$, were captured sequentially without any environment change. 
 Sequences $H_{a_1}^{d_2}$ and $H^{d_2}_{a_2}$, were recorded on the same areas on a different day with object-level scene changes. This dataset is used for both change detection and multi-session mapping evaluation.

 \vspace{0.25em}
 \noindent\textbf{(E) Exo.} This dataset consists of five sequences, $E^{d_1}_{a_{i=1\cdots5}}$, collected with the sensing unit mounted on a self-balancing exoskeleton (\figref{fig:cover}\textbf{b}) teleoperated through a mixed office and lab environment. Each session covers a different portion of the space, with overlapping regions between sessions. This dataset is intended to evaluate LT-Exosense in a real-world exoskeleton deployment scenario.

Ground-truth point cloud maps are provided for all sequences. For the \emph{Human} dataset, ground-truth is obtained using a millimeter-accurate Leica BLK360 terrestrial LiDAR scanner. For the \emph{Exo} dataset, we use a handheld LiDAR-SLAM system~\cite{Ramezani2020Online}. Note that minor background activity occurred during \emph{Exo} recordings, so individual session maps may not perfectly align with the fused ground-truth map.

For all experiments, individual SLAM sessions are generated using an implementation of the Exosense pipeline~\cite{wang2025exosense}, which employs OpenVINS~\cite{Geneva2020OpenVINS} for visual-inertial odometry and LEXIS~\cite{Kassab2024Lexis} to build a pose graph. Point cloud submaps and images are associated with graph vertices and used for subsequent multi-session merging and evaluation.
All processing was performed offline on recorded logs using a mid-range laptop (Intel i7-10750H @ 2.60GHz, 12-core CPU, NVIDIA GTX 1650Ti GPU). All components are CPU-based, except for intrasession visual place recognition, which uses a learning-based model requiring a GPU.

\begin{figure*}
  \centering
  \includegraphics[width=\linewidth]{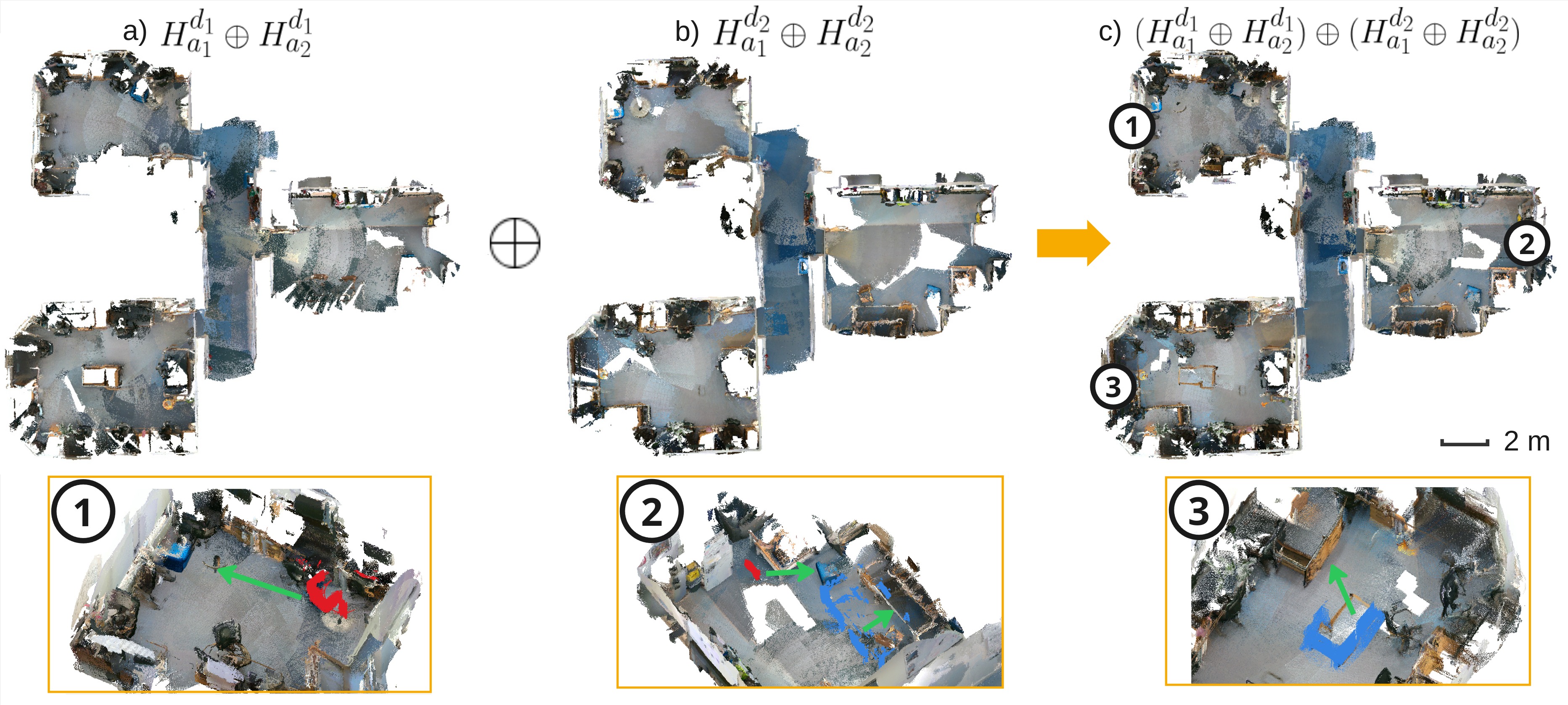}
  \caption{Multi-session mapping and change detection results sequences ${}^{s1}H_a$ and  ${}^{s2}H_b$. Subfigures (a) and (b) show merged maps from sequences recorded on two different days where no environmental change has occurred within each day. Subfigure (c) shows the merged maps from (a) and (b), where detected inter-day object-level change is highlighted in red and blue in the zoomed-in views with green arrows indicating the before and after changes. $\oplus$ here indicates sessions merging followed by change detection and map update.}
  \label{fig:ex_merging}
  \vspace{-10pt}
\end{figure*}

\subsection{Multi-session Trajectory Alignment Accuracy}\label{sec:exp_trajectory}
To evaluate the multi-session trajectory alignment accuracy of LT-Exosense, we compare it against \emph{maplab}~\cite{schneider2018maplab} using a consistent odometry frontend, OpenVINS~\cite{Geneva2020OpenVINS} (referred to as \emph{ov\_maplab}). Both systems are tested on the EuRoC dataset, where multiple sequences are aligned to a common frame and compared against the ground-truth trajectory. We report the Absolute Trajectory Error (ATE) and Relative Pose Error (RPE) of the aligned trajectories in \tabref{tb:trajectory}.

Overall, the multi-session pose graph module of LT-Exosense demonstrates competitive alignment performance. It achieves lower ATE than  \emph{ov\_maplab} but slightly higher RPE across most sequences, indicating strong global consistency but a slight degradation in local accuracy. This trade-off stems from our reliance on a sparse pose graph from an independent SLAM system, which omits high-frequency odometry. This is a current design choice rather than a fundamental limitation, and future work will explore incorporating a denser pose graph to improve local performance.

\input{table_trajectory}

\subsection{Change Detection Performance [\textbf{Human}]}\label{sec:exp_change}
We evaluated LT-Exosense's object-level change detection performance on the \textbf{Human} dataset by merging pairs of sessions that cover the same area but were recorded at different times. 
The output includes both added and removed point clouds, computed using an octree with \SI{5}{\centi\meter} resolution.  For ground truth, we manually annotated the changed regions on the corresponding ground truth scans, and aligned the LT-Exosense outputs accordingly. Since the change detection module also removes noisy or misaligned points arising from session merging errors, we restrict the evaluation to areas that truly contain environment changes.

We evaluated change detection performance as a classification problem. We defined \emph{true positive} (TP) and \emph{false positive} (FP) as the detected changes that are close to the ground truth changed and static points, respectively, while the \emph{false negative} (FN) and \emph{true negative} (TN) are detected static points that are close to the ground truth changed and static points. 
We used the same \SI{5}{\centi\meter} threshold to associate predicted and ground truth changes and report standard classification metrics---precision, recall and F-score values (\tabref{tb:change_detection}). Additionally, we compute the Chamfer Distance between the detected and ground truth changes to quantify the discrepancy of the two point clouds.

Our quantitative results show that LT-Exosense achieves high precision, indicating detected changes have few outliers and align well with actual environmental modifications. However, recall is lower, primarily due to a limited sensor field-of-view and incomplete coverage during traversals, which leads to missed detections when ground-truth areas are unobserved. Despite this, the average Chamfer Distance remains low at \SI{4}{\centi\meter}, suggesting that the spatial reconstruction of detected changes is accurate.

\input{table_change_detection}

\subsection{Multi-session Mapping Quality [\textbf{Human} \& \textbf{Exo}]}\label{sec:exp_mapping}
We next evaluated the reconstruction quality of LT-Exosense using custom sequences under two conditions: \begin{inparaenum}\item merging sessions with no environmental changes, and \item merging sessions that include changes. \end{inparaenum}

For sessions recorded on the same day without change, we ran the LT-Exosense pipeline (with the change detection module disabled) to merge them into a multi-session map, as illustrated in \figref{fig:ex_merging}. Additionally, for the \textbf{Human} dataset, sequences recorded on the same days can be concatenated together and processed as a single SLAM session (denoted as $H_{\text{SLAM}}^{d_1}$ and $H_{\text{SLAM}}^{d_2}$)

For sessions that contain inter-session change, we applied the full LT-Exosense pipeline, including map merging and change detection, to generate updated maps. The quality of the final merged output was evaluated against the ground-truth LiDAR scans (\figref{fig:ex_merging}).

We quantified reconstruction quality using point-to-point distances between the merged map and ground truth (\tabref{tb:mapping}). In the absence of scene change, LT-Exosense achieves mapping accuracy comparable to a single-session SLAM pipeline, demonstrating its ability to incrementally build consistent maps even when not contiguously recorded.

In scenarios involving environmental change, LT-Exosense shows higher maximum point-to-point errors. In the \emph{Exo} dataset, this is primarily due to background activity during recording, which meant that the ground truth LiDAR map was slightly different from the environment---due to people moving during their jobs. In the \emph{Human} dataset, higher error arises when geometry from earlier sessions becomes occluded in later traversals. This happens due to the change in the new session occluding those points. Since these outdated geometries are not explicitly removed unless observed again, they may persist in occluded areas. However, these residual elements typically have minimal impact on downstream navigation, as they are not visible or reachable during the latest traversal.

Since the change-aware map merging pipeline maintains comparable reconstruction accuracy to merging sessions where there is no change detection, this demonstrates that LT-Exosense preserves mapping quality despite scene change.

\input{table_mapping}

\subsection{Path Planning Demonstration}\label{sec:exp_planning}
\begin{figure*}
  \centering
  \includegraphics[width=\linewidth]{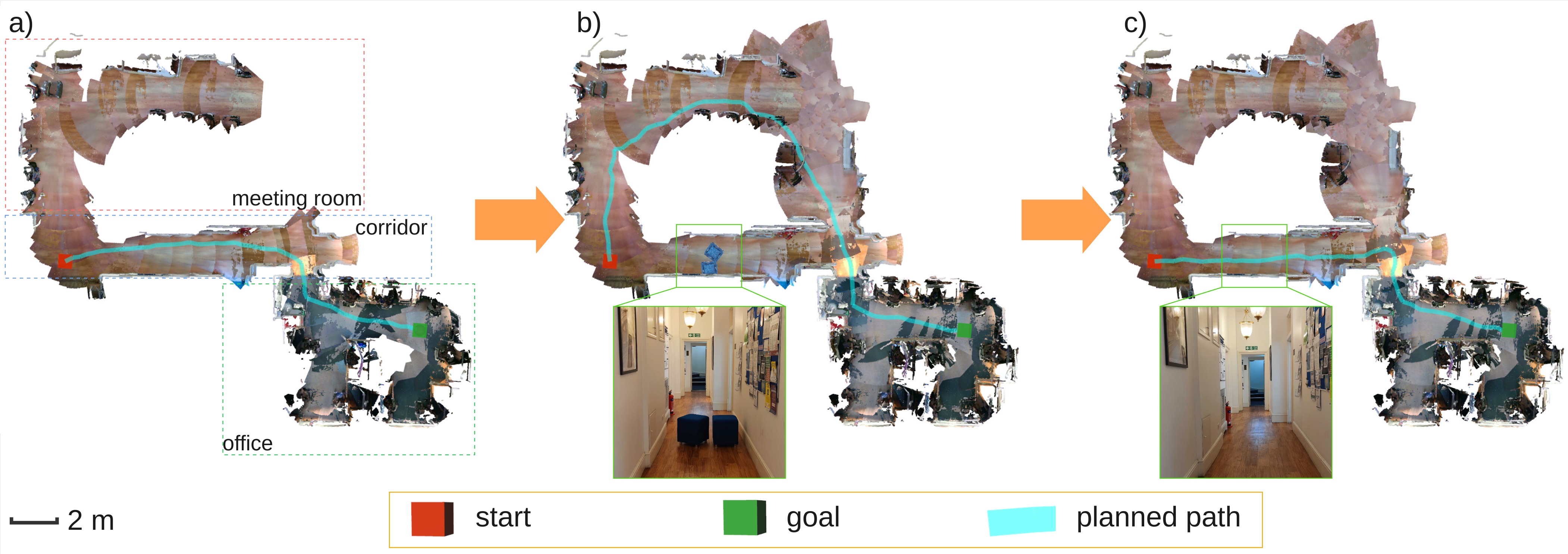}
  \caption{Change-aware adaptive path planning with LT-Exosense through progressive multi-session mapping. From left to right, each subfigure illustrates the planned path from the same start to the same goal as a new mapping session is incrementally merged. \textbf{a)} In Session 1, a direct path through the corridor is planned to reach the goal in the office. \textbf{b)} After merging Session 2, the map becomes more complete but two obstacles appear in the middle of the corridor. LT-Exosense re-plans the path to bypass the obstruction by rerouting through the meeting room. \textbf{c)} When the obstacle is removed in Session 3, the path planner can recover the shorter route again.}
  \label{fig:ex_planning}
  \vspace{-5pt}
\end{figure*}

To qualitatively demonstrate the ability of LT-Exosense's path planning module to adapt to environmental change, we conducted an experiment in a representative indoor environment (\figref{fig:ex_planning}a). To ensure realistic collision modeling during planning, we approximated the physical size of a walking exoskeleton (or human operator) using a bounding box of \si{\numproduct{0.5 x 0.5 x 1.8}}{\unit{\cubic\metre}}.

We designed three mapping sessions, each capturing different regions and the environmental state of the same floor:

\vspace{0.25em}
\noindent\textbf{Session 1.} begins with partial exploration of a meeting room, proceeds to the start of a corridor, traverses through the corridor and enters an office. For the path planning experiment, a path planned from the start pose in the corridor to the goal pose in the office is drawn (\figref{fig:ex_planning}a). 

\vspace{0.25em}
\noindent\textbf{Session 2.} starts in the office and completes the meeting room map by merging with Session 1. Upon traversing the corridor, it discovers a new obstacle blocking the path. Using the same start and goal, the planner then reroutes the robot from the corridor's start, detouring through the newly mapped meeting room to reach the office (\figref{fig:ex_planning}b). (We note again that this system operated passively on recorded logs rather than running live on a robot).

\vspace{0.25em}
\noindent\textbf{Session 3.} remaps the corridor area with the obstacles removed. When Session 3 is merged into the existing map, LT-Exosense correctly identifies the updated changes in the environment and automatically recovers the shorter path for the same start and goal poses as in Session 1 (\figref{fig:ex_planning}c). 

This experiment highlights the incremental map building, map update with change detection, and adaptive path planning capabilities of LT-Exosense. By fusing disjoint exploration sessions, the system forms a coherent, evolving spatial memory. It adapts navigation strategies, choosing longer detours when shortest paths are blocked by change and reverting to optimal paths when obstructions are cleared. 
Ultimately, we envision these capabilities forming the basis of a protective safety system for an exoskeleton user. The system's persistent spatial understanding and reactivity to environment changes would therefore enhance user safety and guidance in dynamic, real-world settings.

\section{Conclusions}\label{sec:conclusion}
We presented \emph{LT-Exosense}, a change-aware, multi-session mapping system for the long-term deployment of self-balancing exoskeletons in evolving environments.
Through experiments, we demonstrated its ability to accurately detect object-level change, maintain high-quality multi-session maps, and support adaptive path planning in dynamic environments. These results position LT-Exosense as a practical system that helps assistive exoskeletons to achieve robust, long-term autonomy.
In future work, we will explore tighter integration of the system with navigation modules in the exoskeleton and its extended deployments in home, rehabilitation, and public environments.

\balance
\bibliographystyle{IEEEtran}
\bibliography{paper}

\end{document}

%% file: preamble.tex
\usepackage{times}
\usepackage{amsmath,amsfonts}
\usepackage{algorithmic}
\usepackage{algorithm}
\usepackage{array}
\usepackage{balance}
\usepackage[dvipsnames,table]{xcolor}
\usepackage{textcomp}
\usepackage{stfloats}
\usepackage{url}
\usepackage{verbatim}
\usepackage{graphicx}
\usepackage{todonotes}
\usepackage{booktabs} %needed for nice tables
\usepackage{bbm} % indicator function
\usepackage{mathtools}
\usepackage{pgf} % for calculating the values for gradient
\usepackage{etoolbox} % robustify command
\usepackage{amssymb}

\usepackage{soul}

\usepackage{paralist}% in-paragraph list
\usepackage{enumitem} % Control indentation of lists

\usepackage{multicol}
\usepackage{hyperref} % to show links to the references, and also back to where the reference was cited
\usepackage{lipsum} % Dummy text (lorem ipsum)
\usepackage{siunitx} % SI units
\usepackage{mathtools}
\usepackage{xspace}

\usepackage{multirow}
\usepackage{pbox}

\usepackage{threeparttable}
\usepackage{tablefootnote}

\usepackage{tikz}

\usepackage[normalem]{ulem}
\usepackage{gensymb}

\definecolor{MK_Two_One}{RGB}{178,24,43} % Nice red
\definecolor{MK_Two_Two}{RGB}{239,138,98}
\definecolor{MK_Two_Three}{RGB}{253,219,199}
\definecolor{MK_Two_Four}{RGB}{209,229,240}
\definecolor{MK_Two_Five}{RGB}{103,169,207}
\definecolor{MK_Two_Six}{RGB}{33,102,172} % Nice blue

\hypersetup{
colorlinks=true
,linkcolor=black
,citecolor=black
,filecolor=MK_Two_Six
,urlcolor= MK_Two_Six
,menucolor=MK_Two_Five
,runcolor=MK_Two_Four
,linkbordercolor=MK_Two_One
,citebordercolor=MK_Two_Two
,filebordercolor=MK_Two_Three
,urlbordercolor=MK_Two_Six
,menubordercolor=MK_Two_Five
,runbordercolor=MK_Two_Four
}

\usepackage{subcaption}
\definecolor{high}{RGB}{116, 173, 209}  % the color for the highest number in your data set
\definecolor{low}{RGB}{244, 109, 67}  % the color for the lowest number in your data set
\def\secref#1{Sec.~\ref{#1}}

\def\figref#1{Fig.~\ref{#1}}
\def\tabref#1{Tab.~\ref{#1}}
\def\eqref#1{Eq.~(\ref{#1})}

\usepackage{tikz}

%% file: glossary.tex
\usepackage[abbreviations]{glossaries-extra}

\glssetcategoryattribute{abbreviation}{indexonlyfirst}{true}

\glssetcategoryattribute{abbreviation}{nohyperfirst}{true}

\newabbreviation{auroc}{AUROC}{Area Under the Receiver Operating Characteristic Curve}
\newabbreviation{accuracy}{Acc}{Accuracy}
\newabbreviation{ate}{ATE}{Absolute Trajectory Error}

\newabbreviation{bev}{BEV}{Bird-Eye View}

\newabbreviation{clip}{CLIP}{Contrastive Language-Image Pretraining}
\newabbreviation{cnn}{CNN}{Convolutional Neural Network}

\newabbreviation{dof}{DoF}{DoF}

\newabbreviation{fov}{FoV}{Field of View}
\newabbreviation{fpr}{FPR}{False Positive Ratio}

\newabbreviation{gnn}{GNN}{Graph Neural Network}
\newabbreviation{gcn}{GCN}{Graph Convolutional Network}

\newabbreviation{knn}{KNN}{K-Nearest Neighbors}

\newabbreviation[plural=LLMs,firstplural=Large Language Models]{llm}{LLM}{Large Language Model}
\newabbreviation{lidar}{LiDAR}{Light Detection and Ranging}

\newabbreviation{mlp}{MLP}{Multi-Layer Perceptron}
\newabbreviation{mpc}{MPC}{Model Predictive Controller}
\newabbreviation{mse}{MSE}{Mean Squared Error}

\newabbreviation{ood}{OOD}{out-of-distribution}

\newabbreviation{pca}{PCA}{Principal Component Analysis}

\newabbreviation{rbf}{RBF}{Radial Basis Function}
\newabbreviation{rmp}{RMP}{Riemannian Motion Policies}
\newabbreviation{rmse}{RMSE}{Root Mean Square Error}
\newabbreviation{ros}{ROS}{Robot Operating System}
\newabbreviation{ros1}{ROS~1}{Robot Operating System}
\newabbreviation{roc}{ROC}{Receiver Operating Characteristic}
\newabbreviation{rpe}{RPE}{Relative Pose Error}
\newabbreviation{rf}{RF}{Random Forest}

\newabbreviation{sdf}{SDF}{Signed Distance Field}
\newabbreviation{slam}{SLAM}{Simultaneous Localization and Mapping}
\newabbreviation{sota}{SOTA}{state-of-the-art}
\newabbreviation{svm}{SVM}{Support Vector Machine}
\newabbreviation{svc}{SVC}{Support Vector Classifier}

\newabbreviation{vit}{ViT}{Vision Transformer}
\newabbreviation{vpr}{VPR}{Visual Place Recognition}
\newabbreviation{vlm}{VLM}{Vision-Language Model}

%% file: notation.tex
\newcommand{\pose}[3]{\mathbf{T}_{\mathtt{#1 #2}}_{#3}}
\newcommand{\rot}[3]{\mathbf{R}_{\mathtt{#1 #2}}_{#3}}
\newcommand{\pos}[2]{{\mathtt{_#1}} \mathbf{p}_{#2}}

\newcommand{\K}{\mathbf{K}_{3\times3}}

\newcommand{\img}[1]{\mathbf{I}^{#1}}

\newcommand{\feat}[1]{\mathbf{F}^{#1}}

\newcommand{\loss}[1]{\mathcal{L}_{\mathrm{#1}}}

\newcommand{\fun}[2]{f_{\mathrm{#1}}\left( #2 \right) }

\robustify{\pose}
\robustify{\rot}
\robustify{\pos}
\robustify{\K}
\robustify{\loss}
\robustify{\feat}
\robustify{\img}
\robustify{\fun}

\definecolor{TraversableBlue}{RGB}{49, 54, 149}
\definecolor{UntraversableRed}{RGB}{192, 26, 38}
\definecolor{PaperOrange}{RGB}{251, 151, 39}
\definecolor{PaperMagenta}{RGB}{150, 36, 145}
\definecolor{PaperBlue}{RGB}{67, 110, 176}
\definecolor{PaperCyan}{RGB}{66, 173, 187}

\usepackage{amssymb}

\usepackage{color}
\usepackage{xcolor}

%% file: table_trajectory.tex
\begin{table}[tbp]
\setlength{\tabcolsep}{3pt}
\centering
\caption{Comparison of multi-session trajectory alignment accuracy between ov\_maplab and LT-Exosense in terms of the root mean squard error (RMSE) of the absolute trajectory error (APE) and relative pose error (RPE).} 
\label{tb:trajectory}
\resizebox{0.95\columnwidth}{!}{%
\begin{tabular}{ccccccccc}
\toprule
\multicolumn{9}{c}{\textbf{Multi-session Trajectory Alignment Accuracy}} \\
\midrule
 \multirow{2}{*}{\shortstack{Dataset}} & \multirow{2}{*}{\shortstack{Seq.}}  & \multicolumn{3}{c}{ov\_maplab} & \multicolumn{3}{c}{LT-Exosense} & \multirow{3}{*}{\shortstack{Length\\ (m)}} \\
& &\multicolumn{1}{c}{ATE}  & \multicolumn{1}{c}{RPE (\SI{1}{\meter})} &\multirow{1}{*}{\shortstack{No. Poses}}  &\multicolumn{1}{c}{ATE}  & \multicolumn{1}{c}{RPE (\SI{1}{\meter})} & \multirow{1}{*}{\shortstack{No. Poses}}  & \\[5pt]
\midrule

\multirow{3}{*}{\shortstack{V1}} & V1\_01 & 0.063 & 0.086 & 2774 & 0.056 & 0.107 & 126 & 58.6 \\  
&  V1\_02 & 0.057 & 0.034 & 1598 & 0.043 & 0.049 & 78 & 75.9 \\
& V1\_03 & 0.06 & 0.04 & 1988 & 0.08 & 0.053 & 97 & 79 \\
\midrule
\multirow{3}{*}{\shortstack{V2}} & V2\_01 & 0.059 & 0.033 & 2170 & 0.04 & 0.044 & 94 & 36.5 \\  
&  V2\_02 & 0.045 & 0.024 & 2234 & 0.053 & 0.03  & 112 & 83.2 \\
& V2\_03 & 0.103 & 0.038 & 1766 & 0.086 & 0.048  & 108 & 86.1 \\
\bottomrule
\end{tabular}
} % end of the resizebox command
\vspace{-5pt}
\end{table}

%% file: table_change_detection.tex
\begin{table}[tbp]
\setlength{\tabcolsep}{3pt}
\centering
\caption{Change detection performance metrics. For every two sessions of the same area but recorded at different times, we merge them and perform change detection  in both directions.} 
\label{tb:change_detection}
\resizebox{0.9\columnwidth}{!}{%
\begin{tabular}{ccccc}
\toprule
\multicolumn{5}{c}{\textbf{Change Dectection Evaluation}} \\
\midrule
Comp.  & Precision & Recall  & F-score  & Chamfer Dist. [m]\\
\midrule
$H^{d_1}_{a_1} \rightarrow H^{d_2}_{a_1}$ &\SI{88.9}{\percent} &  \SI{41.9}{\percent} & \SI{57.0}{\percent} & 0.042 \\
 $H^{d_2}_{a_1} \rightarrow H^{d_1}_{a_1}$ & \SI{88.4}{\percent} &  \SI{48.4}{\percent} & \SI{62.5}{\percent} & 0.034 \\
$H^{d_1}_{a_2} \rightarrow H^{d_2}_{a_2}$ & \SI{90.9}{\percent} &  \SI{53.4}{\percent} & \SI{67.3}{\percent} & 0.043 \\
 $H^{d_2}_{a_2} \rightarrow H^{d_1}_{a_2}$ & \SI{80.1}{\percent} &  \SI{35.7}{\percent} & \SI{49.4}{\percent} & 0.04 \\

\bottomrule
\end{tabular}
} % end of the resizebox command
\end{table}

%% file: table_mapping.tex
\begin{table}[tbp]
\setlength{\tabcolsep}{3pt}
\centering
\caption{Multi-session Mapping Quality. Using the point-to-point distance between the estimated map against the ground truth scan, we compute the mean, median, max and $90^{\text{th}}$ percentile error to quantify the  multi-session mapping quality. We use $\oplus$ operator to denote multi-session merging operation.} 
\label{tb:mapping}
\resizebox{0.95\columnwidth}{!}{%
\begin{tabular}{ccccc}
\toprule
\multicolumn{5}{c}{\textbf{Point-to-point Distance of Multi-session Mapping}} \\
\midrule
Seq.  & mean [m] & median [m]  & max [m]  & 90\% [m] \\
\midrule
$H^{d_1}_{\text{SLAM}}$ & 0.024 & 0.017 & 0.264 &  0.05 \\
$H^{d_1}_{a_1} \oplus H^{d_1}_{a_2}$ & 0.028 & 0.016 &  0.325 & 0.063 \\
\midrule
$H^{d_2}_{\text{SLAM}}$ & 0.031 &	0.019 & 0.47 & 0.067 \\
$H^{d_2}_{a_1} \oplus H^{d_2}_{a_2}$ & 0.031 &  0.021 & 0.46 & 0.068 \\
\midrule
$(H^{d_1}_{a_1} \oplus H^{d_1}_{a_2}) \oplus (H^{d_2}_{a_1} \oplus H^{d_2}_{a_2}) $  & 0.032 & 0.022 & 0.75 &	0.069 \\
\midrule
$E^{d_1}_{a_1} \oplus E^{d_1}_{a_2} \oplus E^{d_1}_{a_3}  \oplus E^{d_1}_{a_4}  \oplus E^{d_1}_{a_5} $ & 0.046 & 0.027 & 0.873 & 0.099 \\ 
\bottomrule
\end{tabular}
} % end of the resizebox command
\end{table}